\documentclass{article}

\usepackage[final]{corl_2026} %
\PassOptionsToPackage{dvipsnames,table}{xcolor}
\usepackage{xcolor}
\usepackage{corl_2026} %

\usepackage{amsmath,amssymb,amsthm}
\usepackage{graphicx}
\usepackage{tabularx}
\usepackage{booktabs}
\usepackage{makecell}
\usepackage{enumitem}
\usepackage{caption} 
\usepackage{adjustbox}
\usepackage{float}
\usepackage{tikz}
\usetikzlibrary{decorations.pathmorphing, arrows.meta, decorations.pathreplacing}

 \newcommand{\apref}[1]{Appendix~\ref{#1}} %

\theoremstyle{definition}

\theoremstyle{remark}

\newcommand{\E}{\mathbb{E}}

\newcommand{\baseline}[1]{\texttt{#1}}
\newcommand{\scenario}[1]{\textsc{#1}}

\usepackage{enumitem}
\setlist{itemsep=2pt, topsep=2pt, parsep=0pt, partopsep=0pt}

\usepackage{xspace}

\newcommand{\paveNaiveSFMax}{48.5~pp\xspace}
\newcommand{\paveNaiveSFMean}{27.9~pp\xspace}

\makeatletter
\@ifpackageloaded{enumitem}{}{\usepackage{enumitem}}
\@ifpackageloaded{etoolbox}{}{\usepackage{etoolbox}}

\newcounter{rbissue}
\renewcommand{\therbissue}{Q\number\c@rbissue}

\newcommand{\rb@register}[1]{%
  \begingroup\edef\rb@tmp{\endgroup
    \noexpand\listcsgadd{rbidx@#1}{\therbissue}}\rb@tmp
}

\newcommand{\rbIssue}[3]{%
  \refstepcounter{rbissue}%
  \forcsvlist{\rb@register}{#3}%
  \item[\textbf{\therbissue:}]\label{#2}%
  \textbf{#1}~{\normalfont(#3)}\par\nobreak\ignorespaces
}

\newcommand{\rb@printtag}[1]{\rb@sep #1\def\rb@sep{, }}
\newcommand{\rbIndexFor}[1]{%
  \ifcsundef{rbidx@#1}{\textit{none}}%
    {\def\rb@sep{}\forlistcsloop{\rb@printtag}{rbidx@#1}}%
}
\makeatother

\title{Leave It Better Than Found: Value-Guided Cooperation in Sequential Tasks}

\author{
  Mario Gomez Andreu$^{1,2}$ \\
  \texttt{margomez@ethz.ch}
  \And
  Michael Amir$^{1}$ \\
  \texttt{ma2151@cam.ac.uk}
  \And
  Antonio Marino$^{1}$ \\
  \texttt{am3507@cam.ac.uk}
  \And
  Tianxu An$^{2}$ \\
  \texttt{tianan@ethz.ch}
  \And
  Marco Hutter$^{2}$ \\
  \texttt{mahutter@ethz.ch}
  \And
  Amanda Prorok$^{1}$ \\
  \texttt{asp45@cam.ac.uk}
}

\begin{document}
\maketitle

{\let\thefootnote\relax\footnotetext{%
  $^{1}$Prorok Lab, Department of Computer Science and Technology, University of Cambridge, Cambridge, United Kingdom 
  $^{2}$Robotic Systems Lab, Department of Mechanical and Process Engineering, ETH Z\"urich, Z\"urich, Switzerland}}

\definecolor{leadercol}{HTML}{E65100} %
\definecolor{followcol}{HTML}{1565C0} %
\definecolor{dirtcol}{HTML}{D32F2F}   %
\definecolor{vlow}{HTML}{440154}      %
\definecolor{vmid}{HTML}{21908C}      %
\definecolor{vhigh}{HTML}{FDE725}     %

\begin{abstract}

We consider the problem of \textit{sequential task continuation}, where one robot acts in a workspace that has been affected by a previous robot.  
Current approaches in multi-robot cooperation rely heavily on co-training or co-optimization, assuming that the robots have access to collective interaction data during the policy training or planning process. However, as increasingly varied robotic embodiments and policies are deployed and operate in the same real-world environment, learning policies for cooperation becomes infeasible due to the combinatorial explosion of possible robot combinations that need to be accounted for. To address this challenge, this paper considers the problem of \textit{ad hoc} sequential cooperation between robots that have never before operated together in a shared environment. We demonstrate that robots can explicitly communicate their environmental preferences through sharing their \textit{value functions}, enabling coordination without the need for prior joint training. To ground this exchange and prevent overconfidence in out-of-distribution states, we further introduce a $k$-nearest neighbors (KNN)-based uncertainty filter. Our method yields an increase of up to \paveNaiveSFMax compared with naive or even selfish agents, in task success rates across simulated and real-world experiments, demonstrating that value function transmission between agents is useful not merely as an internal learning tool, but as an effective basis for ad hoc coordination. \\
\textbf{Website:} \href{https://proroklab.github.io/PAVE/}{https://proroklab.github.io/PAVE/}
\end{abstract}

\begin{figure}
    \centering
    \begin{adjustbox}{minipage=1.0\textwidth, center}
    \begin{minipage}[c]{0.85\linewidth}
        \begin{minipage}[c]{0.1\linewidth}~\end{minipage}\hfill
        \begin{minipage}[c]{0.9\linewidth}
            \centering
            \begin{minipage}[c]{0.33\linewidth}\centering\textit{Initial Condition}\end{minipage}\hfill
            \begin{minipage}[c]{0.33\linewidth}\centering\textit{Leader: Clean under table }\end{minipage}\hfill
            \begin{minipage}[c]{0.33\linewidth}\centering\textit{Follower: Traverse room}\end{minipage}
        \end{minipage}\\[4pt]

        \begin{minipage}[c]{0.1\linewidth}
            \centering
            \rotatebox{90}{\textbf{Naive}}
        \end{minipage}\hfill
        \begin{minipage}[c]{0.9\linewidth}
            \centering
            \includegraphics[width=0.33\linewidth, keepaspectratio]{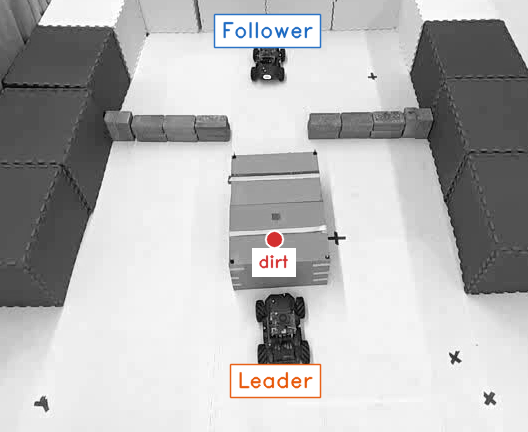}\hfill
            \includegraphics[width=0.33\linewidth, keepaspectratio]{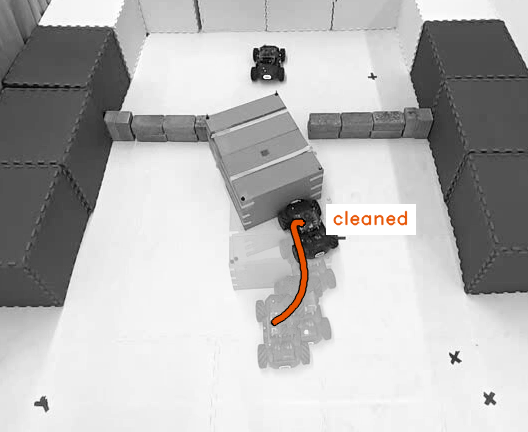}\hfill
            \includegraphics[width=0.33\linewidth, keepaspectratio]{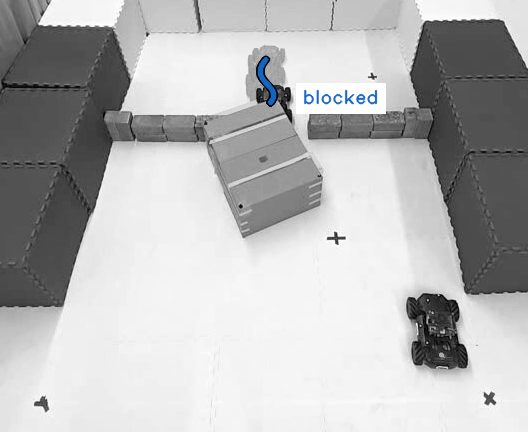}\\[4pt]
        \end{minipage}\\[1pt]
          \begin{minipage}[c]{0.1\linewidth}~\end{minipage}\hfill
         \begin{minipage}[c]{0.9\linewidth}
            \centering
            \begin{tikzpicture}
                \draw[->, line width=3pt, black!40] (0,0) -- (\linewidth-1.2cm,0)
                    node[midway, above, font=\small\bfseries, black!60] {Time};
            \end{tikzpicture}
        \end{minipage}\\[4pt]
        \begin{minipage}[c]{0.1\linewidth}
            \centering
            \rotatebox{90}{\textbf{PAVE (Ours)}}
        \end{minipage}\hfill
        \begin{minipage}[c]{0.9\linewidth}
            \centering
            \includegraphics[width=0.33\linewidth, keepaspectratio]{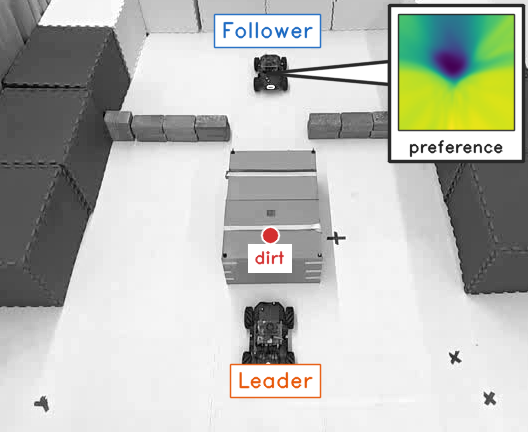}\hfill
            \includegraphics[width=0.33\linewidth, keepaspectratio]{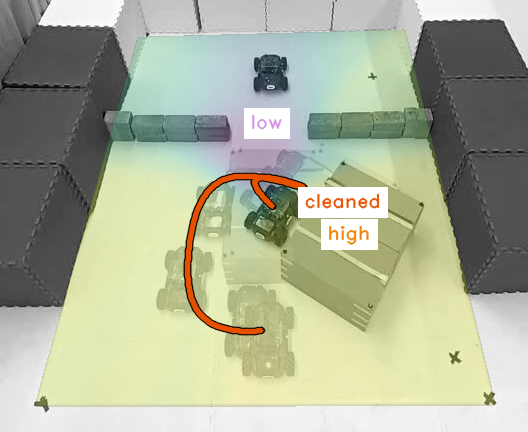}\hfill
            \includegraphics[width=0.33\linewidth, keepaspectratio]{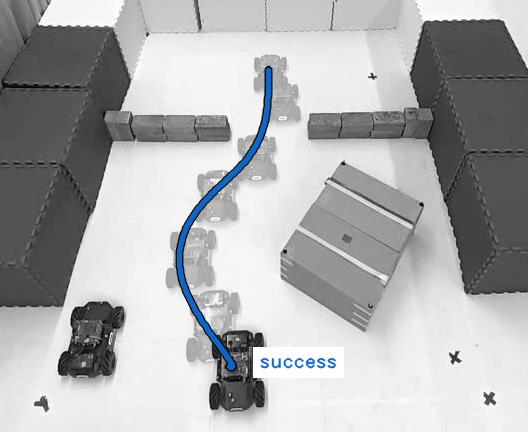}
        \end{minipage}
    \end{minipage}\hfill
\begin{minipage}[c]{0.135\linewidth}
        \centering
        {\footnotesize\textbf{Legend}}\\[4pt]
        {\scriptsize\setlength{\tabcolsep}{2pt}%
        \begin{tabular}{@{}c@{\ \ }p{0.72\linewidth}@{}}
        \tikz[baseline=-0.5ex]{\draw[leadercol,line width=1.4pt](0,0)--(0.4,0);} &
            \raggedright Leader trajectory \tabularnewline[4pt]
        \tikz[baseline=-0.5ex]{\draw[followcol,line width=1.4pt](0,0)--(0.4,0);} &
            \raggedright Follower trajectory \tabularnewline[4pt]
        \tikz[baseline=-0.5ex]{\shade[left color=vlow,right color=vmid](0,0)rectangle(0.22,0.14);\shade[left color=vmid,right color=vhigh](0.22,0)rectangle(0.44,0.14);\draw[black!40,line width=0.2pt](0,0)rectangle(0.44,0.14);} &
            \raggedright Preference for table position (low\,$\rightarrow$\,high) \tabularnewline
        \end{tabular}}
    \end{minipage}
    \end{adjustbox}
    \caption{Two agents perform tasks sequentially in a shared environment: a leader pushes the table aside to clean beneath it; a follower must then traverse the room. \textbf{Naive} (top), the leader is unaware of the follower's needs and leaves the table blocking the path, preventing traversal. \textbf{PAVE (Ours)} (bottom), the follower shares its value function over environment states, and the leader adapts by placing the table in a high-value region (overlaid with heatmap) for the follower, clearing the path, and enabling traversal.}
    \label{fig:hero}
\end{figure}

\section{Introduction}
\label{sec:introduction}

As robotic systems achieve greater autonomy, heterogeneous agents, e.g., cleaners, delivery bots, and security systems, will increasingly operate in the same public and domestic spaces. Robots modify these spaces to achieve their goals, leaving subsequent actors with the environments that they left behind.
These workspace overlaps require spontaneous cooperation between independently developed systems. We formalize this challenge as \textit{sequential task continuation}: a setting where an agent must proactively anticipate how its physical actions will alter the shared space and affect subsequent actors.

Currently, robot behaviors are typically designed in isolation by different parties to perform distinct tasks. When actors interact in the same shared workspace, we require methods and protocols that bridge these independent systems. For instance, delivery bots should not drop packages along common patrol routes, and cleaners should not rearrange the environment in such a way that areas become inaccessible to other robots. Because hand-crafted rules are too brittle to handle the infinite edge cases of open-world environments, learning-based coordination is required. Training a separate cooperative policy for every possible pairing of robots is unrealistic, as the possible combinations increase exponentially with the number of robot embodiments and tasks.

 Instead, we turn to \textit{ad hoc} coordination between agents that does not require joint training, as robots can easily share key components of their decision-making with other agents. Given the representation of other agents' preferences, robots can adapt their behavior on the fly (ad hoc) to aid other robots in their tasks while still pursuing their own. 
 Existing frameworks in this domain, however, rely on restrictive assumptions. Zero-shot coordination \cite{hu_other-play_2020} trains a single robust policy, but it does not allow online adaptation. Ad Hoc teamwork, first proposed in Stone et al.~\cite{stone_ad_2010} and implemented in PLASTIC~\cite{barrett_making_2017} and TEAMSTER~\cite{ribeiro_teamster_2023}, adapts online but assumes a single common payoff and infers teammate behavior from observation rather than direct communication. Both approaches assume that actors act simultaneously and not sequentially.
 
 Our approach is different: agents pursue distinct tasks and exchange compact representations of their preferences. Rather than relying on task-specific learned communication channels~\cite{foerster_learning_2016, li_graph_2020}, we leverage a quantity that reinforcement-learning agents already learn: the value function, which maps states to expected return.
For \textit{sequential task continuation}, the relevant part is the follower's value over shared environment states. Sharing the value function tells the leader which handoff states are useful for the follower.
Our method, \textbf{Policy Adaptation via Value Exchange} (PAVE), uses this single exchange of the value function to coordinate with the following agent without joint training, adapting the leading agent's policy online using a receding horizon planner.
 
\noindent We contribute:
\begin{enumerate}
    \itemsep0em 
    \item The formalization of \textit{sequential task continuation} with environment handoff.
    \item \baseline{PAVE}: A method that combines value-function sharing with an online receding-horizon planner to coordinate two independently trained robots.
    \item An uncertainty-based grounding using a $k$-nearest-neighbors distribution filter to prevent the receiving agent from acting on unreliable value estimates.
  
\end{enumerate}
We empirically show an average \paveNaiveSFMean improvement in follower task success over a non-cooperative baseline (up to \paveNaiveSFMax in the most-constrained case) across five simulated scenarios and a real-world deployment.

\paragraph{Related Work}
Multi-agent reinforcement learning typically relies on joint training with shared critics, as in MAPPO~\cite{yu_surprising_2022}, MADDPG~\cite{lowe_multi-agent_2017}, COMA~\cite{foerster_counterfactual_2018}, and QMIX~\cite{rashid_qmix_2018}. These approaches do not scale to \textit{sequential task coordination} as the robots act not simultaneously but sequentially in a shared environment. Cooperative Consecutive Policies \cite{erskine_developing_2022} instead train distinct policies acting on a shared embodiment to solve long-horizon tasks, by optimizing each current policy with respect to the next agent’s Q-function to ensure an optimal hand-off. Although distinct from our method, it assumes a shared embodiment and joint training, and it inspires one of our baselines.   
Dhakal et al. \cite{dhakal2026} propose a planner for rearrangement tasks that anticipates the effects of its actions on future tasks of the same agent. PAVE differs by reusing an existing value function, filtering out-of-distribution value estimates, and considering multiple agents instead of a single agent.
 
Other approaches similarly treat the environment as the medium of inter-agent influence. Stigmergy \cite{holland_stigmergy_1999, dorigo_ant_2000, rappel_stigmergy-based_2023} is a natural process used by swarm animals such as ants to coordinate through pheromone deposits in the environment. Environment Co-Design~\cite{gao_constrained_2023, gao_co-optimizing_2025, li_scaling_2025} also considers adapting the environment to benefit the agents acting within it. PAVE defines a middle ground: agents communicate directly via their value functions, but enact that coordination indirectly. One agent manipulates the physical space via its task-constrained policy to favorably shape the state for the next agent, achieving continuation without requiring central design or shared training. For an extended discussion of the related literature, we refer the reader to \apref{ap:extended_related_work}.

\section{Problem Statement}
\label{sec:problem_def}

Consider the \scenario{household} scenario in Figure \ref{fig:hero}. A cleaning robot has to move a table to clean below it. Unaware of other actors in the environment, it might reposition the table in front of a pathway, blocking another robot (that does not have the ability to move the table) from subsequently traversing the room.
If the following agents' preferences for the environment are considered, the cleaning robot can still succeed while removing the table from the traversing agent's path and \textit{leaving the environment better than it found it}.

\textbf{Setting.} More generally, we consider cooperation problems in which agents act sequentially in the same environment. We call the first agent the leader $L$ and the second agent the follower $F$. Both share a common representation of their environment $\mathbf{w}$ as well as individual private states $\mathbf{s}^L$ and $\mathbf{s}^F$. They act according to policies $\pi_L$ and $\pi_F$ with reward functions $r_L, r_F$. Their expected returns are $J_L(\pi_L) = \E_{\mathbf{s}^L_0 \sim \rho_L,\, \mathbf{w}_0 \sim \rho_W}\!\left[\sum_{t=0}^{T} \gamma^t \, r_L\big(\mathbf{s}^L_t,\mathbf{w}_t,\pi_L(\mathbf{s}^L_t,\mathbf{w}_t)\big)\right]$, $J_F(\pi_F, \mathbf{w}) = \E_{\mathbf{s}^F_0 \sim \rho_F}\!\left[\sum_{t=0}^{K} \gamma^t \, r_F\big(\mathbf{s}^F_t,\mathbf{w}_t,\pi_F(\mathbf{s}^F_t,\mathbf{w}_t)\big) \,\big|\, \mathbf{w}_0 = \mathbf{w}\right]$ 
where $\rho_L$ and $\rho_F$ are the initial distributions of the leader's and follower's private states, respectively, $\rho_W$ is the initial distribution of the environment state, and $\gamma \in (0,1]$ is the discount factor. The follower's expectation conditions on $\mathbf{w}_0 = \mathbf{w}$, i.e.\ the handoff state passed in from the leader.

\textbf{Sequential Subtask Execution.} The leader acts in the environment for $T$ steps leaving behind a terminal environment state $\mathbf{w} \sim H_T(\pi_L)$ distributed according to the handoff distribution $H_T$, which depends only on the leader policy $\pi_L$ and world dynamics.

\textbf{Cooperative Leader.} To achieve meaningful cooperation, the leader cannot strictly maximize its own return, $J_L$. It must concurrently optimize for the follower's expected return given the inherited environment state. Because the leader's influence on the follower is solely dictated by the handoff distribution, $H_T$, the joint cooperative objective becomes
\begin{equation}
  \max_{\pi_L} \; J_L(\pi_L) + \alpha\, \E_{\mathbf{w} \sim H_T(\pi_L)}\!\left[J_F(\pi_F, \mathbf{w})\right].
  \label{eq:final-opt}
\end{equation}
with weight $\alpha$.
The follower policy is assumed to be fixed, reacting only to the handoff state that the leader leaves behind.

\textbf{Modular Composition} As this work focuses on ad hoc coordination, we assume that collective interaction data is not accessible prior to deployment. This requires the leader to solve ~\eqref{eq:final-opt} online during deployment. Further, the leader needs to evaluate the follower's performance $J_F$ under arbitrary environment states $\mathbf{w}_T$, requiring detailed knowledge about the follower's behavior, its interaction dynamics with the environment, and its reward function to simulate its response to the leader's behavior when planning. We address this problem in the following section by introducing a simplified preference interface that abstracts this complexity and is easy to obtain.

\section{Method}
Modularity stems from abstraction through common interfaces. We propose using the value function as an abstraction for coordination. The value function of a policy maps states to expected returns under that policy. Because agents in a shared environment operate on a common environment state $\mathbf{w}$, the \textit{environmental value function} $V_F(\mathbf{w})$ naturally serves as a preference function, communicating how desirable a given environment state is to the follower. By construction, $V_F(\mathbf{w}) \doteq J_F(\pi_F, \mathbf{w})$, meaning it equals the follower's expected return from the handoff state $\mathbf{w}$.
Substituting into \eqref{eq:final-opt}, the cooperative objective becomes
\begin{equation}
    \max_{\pi_L} \; J_L(\pi_L) + \alpha\, \E_{\mathbf{w} \sim H_T(\pi_L)}\!\left[V_F(\mathbf{w})\right].
    \label{eq:final-opt-value}
\end{equation}
By revealing its value function, the follower enables the leader to optimize \eqref{eq:final-opt-value} on its behalf without requiring any further input.
In practice, we obtain the environmental value function by aggregating the critic of an agent trained with PPO \cite{schulman_proximal_2017}. The critic $V_F^{\mathrm{critic}}(\mathbf{s}, \mathbf{w})$ depends on the full follower state $(\mathbf{s}, \mathbf{w})$, comprising the follower's private state $\mathbf{s}$ and the environmental state $\mathbf{w}$. Since the leader optimizes only over $\mathbf{w}$, we aggregate the critic to produce the environmental value function --- either by fixing the initial private state to a known value $\mathbf{s}_0$ ($V_F(\mathbf{w}) = V_F^{\mathrm{critic}}(\mathbf{s}_0, \mathbf{w})$), or by taking the expectation over the follower's initial private state distribution $V_F(\mathbf{w}) = \E_{\mathbf{s}_0 \sim \rho_F}\!\left[V_F^{\mathrm{critic}}(\mathbf{s}_0, \mathbf{w})\right]$. In this work, we only consider the case of a known initial state for the follower. 
Because PAVE shares only $V_F$, the leader cannot recover $\pi_F$ without an additional model of the follower's private state and dynamics. Specifically, $V_F$ does not expose the follower's individual state or action distribution. 

\paragraph{Cooperative Objective}
Combining two qualitatively different objectives is nontrivial and domain-dependent. The proposed approach allows for the selection of arbitrary aggregation functions.
In this work, we combine the two objectives by first normalizing them and then summing them using a weighting factor, $\alpha$.
\begin{equation}
  	\max_{\pi_L}\; \frac{J_L - J^{min}_L}{\Delta{J_L}} + \alpha \frac{V_F - J^{min}_F}{\Delta{J_F}}
\end{equation}
	
where $J^{min}_L, \Delta{J_L},J^{min}_F,\Delta{J_F}$ are determined from rollouts of the leaders' and followers' prior policies.

\paragraph{Uncertainty-Aware Value Interface}
\label{sec:uncertainty}
Because the leader and follower are not trained jointly, the environment states generated by the leader may shift away from those seen during the follower's training. This distribution shift substantially reduces the accuracy of out-of-distribution value estimates. It is therefore necessary to estimate value uncertainty and adapt the leader’s objective accordingly.  Standard uncertainty quantification methods typically require modifying the follower’s architecture or training pipeline, which violates the decoupled nature of ad hoc cooperation. Instead, we adopt a heuristic that can be evaluated directly from the follower’s training-state distribution: a KNN-based approach to determine whether a handoff state is out of distribution, similar to \cite{sun_out--distribution_2022}:
\begin{equation}
  		\beta(\mathbf{w}_T) = \mathtt{clip}\!\left(0,\,1,\, \frac{\frac{1}{K}\sum_{k=1}^K
  		||\mathtt{neighbor}(\mathbf{w}_T, k) - \mathbf{w}_T||_2 - \tau}{ d_{max} - \tau} \right)
	\end{equation}
$\beta(\mathbf{w}_T)$ is close to zero when environment states are near the training samples on which the follower was trained, and becomes large when they are far away. This can be leveraged to adjust the value function in regions of the state space where the agent, and therefore its value function, has not been trained.
We choose $K = 5$. The parameters $\tau$ and $d_{max}$ are chosen to be, respectively, the 95\textsuperscript{th} and 99.9\textsuperscript{th} percentiles of the empirical distribution of the mean distance to the $K$ nearest neighbors of each state, computed over a sample of 1000 states drawn from the environmental state distribution on which the follower was trained.

\paragraph{Cooperative Online Planning}
\label{sec:planning}
To enable ad hoc cooperation, the planning procedure must be computationally feasible online.  Because trajectory optimization benefits significantly from a strong prior, we use the leader's independently trained policy $\pi_L$ to warm start the search. At each step, the planner seeks to maximize the normalized, uncertainty-aware cooperative objective:
\begin{equation}
    \label{eq:pave_optim_problem}
  	\max_{\pi_L}\; \frac{J_L - J^{min}_L}{\Delta J_L} + \alpha (1 - \beta(\mathbf{w}_{t+H})) \frac{V_F(\mathbf{w}_{t+H}) - J^{min}_F}{\Delta J_F}
\end{equation}
We execute the planner in a receding-horizon fashion at a given time $t$, planning over a horizon of $H$ time steps and predicting future states $\mathbf{s}_{i}, \mathbf{w}_{i}$ and rewards $r^L_i$ using a zero-order model of the system dynamics and reward, as our method does not rely on gradient-based optimization. Because the time horizon is limited, we must estimate the leader's full return, $J_L$, by approximating the tail rewards.
\begin{equation}
	J_L \approx \underbrace{\sum_{i = 1} ^{t-1} \gamma^i r^L_i}_{\text{already incurred}} + \underbrace{\sum_{i = t} ^{t+H} \gamma^i r^L_i}_{\text{within optimization horizon}} + \;
    \underbrace{\gamma^{t+H}\; V_L(\mathbf{s}_{t+H}, \mathbf{w}_{t+H})}_{\text{tail estimation using leader's value function}}
\end{equation}
This limited horizon requires two practical approximations. First, because the true terminal handoff state $\mathbf{w}_T$ is unavailable at time $t$, we use the environment state at the end of the horizon, $\mathbf{w}_{t+H}$, as a proxy to evaluate both the follower's value function $V_F$ and the uncertainty penalty $\beta$. Second, using $V_L$ as an infinite-horizon value approximation introduces a non-cooperative baseline bias: $V_L$ assumes the leader will act purely selfishly beyond the horizon, yielding an overestimate of the true return, since the online policy may actually continue sacrificing some reward to aid the follower. Empirically, however, neither the state proxy nor the terminal value approximation adversely impacts the performance of either agent.
To solve the optimization problem \eqref{eq:pave_optim_problem} online, \baseline{PAVE} employs the Cross Entropy Method (CEM) \cite{rubinstein_cross-entropy_1999} over the receding horizon. 
A detailed listing of parameter values is provided in \apref{ap:planner_parameters}.

\textit{\textbf{Remark:}} This methodology extends to arbitrary sequences of $N > 2$ agents through recursive pairwise application. For an ordered sequence of agents $(1, \dots, N)$, each agent $i \in \{1, \dots, N-1\}$ treats agent $i+1$ as its immediate follower, adapting its policy to the successor's value function $V_{i+1}$. Provided that the two-agent formulation guarantees the leader preserves the follower's baseline performance, this framework can be applied transitively to guarantee cooperative behavior across the entire $N$-agent chain. 

\section{Experimental Analysis and Real-world Deployment}
\begin{figure}[t]
    
    \centering
    \begin{minipage}[t]{0.3\textwidth}
        \centering
        \textbf{\scenario{Warehouse}}\\[2pt]
        \includegraphics[width=0.8\textwidth]{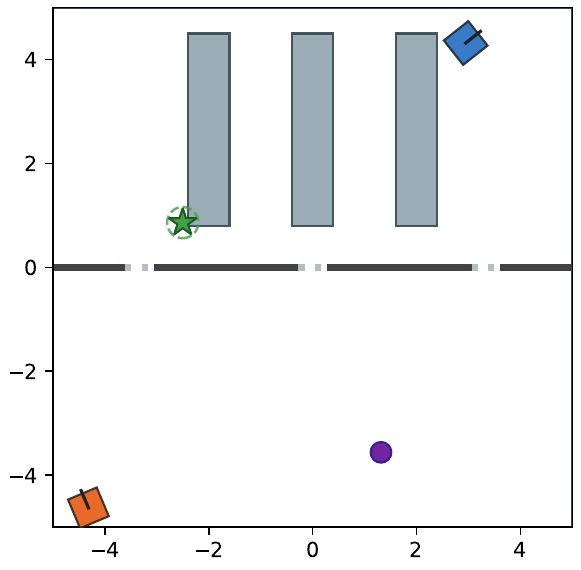}
    \end{minipage}\hfill
    \begin{minipage}[t]{0.3\textwidth}
        \centering
        \textbf{\scenario{Mobile Arm}}\\[2pt]
        \begin{tikzpicture}
            \node[anchor=south west, inner sep=0] (leader) at (0,0)
                {\includegraphics[width=0.8\linewidth]{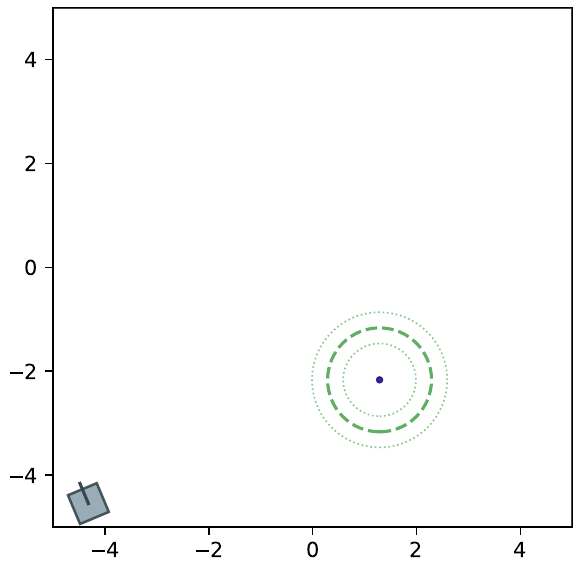}};
            \node[anchor=north east, inner sep=0]
                at (leader.north east)
                {\includegraphics[width=0.35\linewidth]{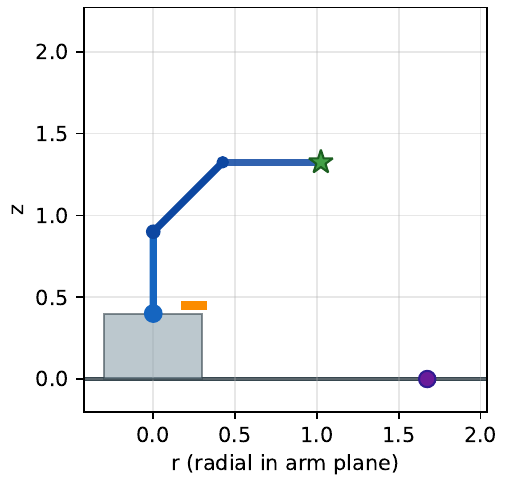}};
        \end{tikzpicture}
    \end{minipage}\hfill
    \begin{minipage}[t]{0.3\textwidth}
        \centering
        \textbf{\scenario{Obstacle Arena}}\\[2pt]
        \includegraphics[width=0.8\textwidth]{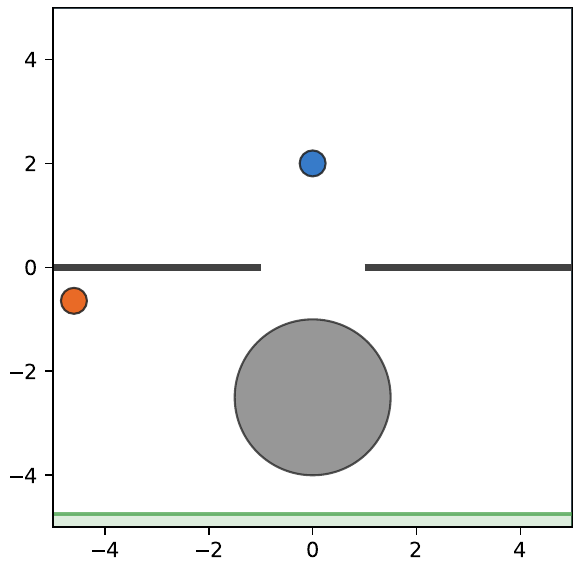}
    \end{minipage}
    \caption{The three simulated scenarios. In each, a leader (first agent) reshapes the shared environment before a follower (second agent) must complete its own task from the resulting handoff state.}
    \label{fig:scenario_cards}
\end{figure}

We evaluate our method in five scenarios --- \scenario{Warehouse}, \scenario{Warehouse Pickup Blocked}, \scenario{Warehouse Pickup Unseen}, \scenario{Mobile Arm}, and \scenario{Obstacle Arena} (Figure \ref{fig:scenario_cards}) --- in simulation, as well as in one real-world experiment (\scenario{Household}). For a description of each scenario,  we refer the reader to Table \ref{tab:scenarios}. Hardware experiments are performed using the RoboMaster Platform \cite{blumenkamp2024robomaster}. Complete details of the simulated environments and the hardware setup are provided in \apref{ap:simulation_environments} and \apref{ap:hw_setup}, respectively.
\begin{table}[ht]
\scriptsize
\centering

\resizebox{\textwidth}{!}{
\begin{tabular}{@{}>{\raggedright\arraybackslash}p{0.16\linewidth}>{\raggedright\arraybackslash}p{0.26\linewidth}>{\raggedright\arraybackslash}p{0.26\linewidth}>{\raggedright\arraybackslash}p{0.26\linewidth}@{}}
\toprule
\textbf{Scenario} & \textbf{Setting} & \textbf{Leader Task} & \textbf{Follower Task} \\
\midrule

\scenario{Mobile Arm}
  & Heterogeneous embodiment: a robotic manipulator mounted on a mobile base.
  & Control the mobile base to navigate to a target object.
  & Control the manipulator to grasp the object; reach is restricted to a cone in front of the base. \\
\hline
\addlinespace
\scenario{Obstacle Arena}
  & A pusher and a navigator share a room with a bottleneck.
  & Push an obstacle out of the center of the room toward the nearest wall.
  & Traverse the bottleneck and reach the opposite wall. \\
\hline
\addlinespace
\scenario{Warehouse}
  & Two agents in a warehouse with three transfer areas (left, center, right).
  & Pick up a randomly placed object and transport it to one of the transfer areas.
  & Retrieve the object from the transfer area and place it at a designated shelf location. \\
\hline
\addlinespace
\scenario{Warehouse (Pickup-blocked)}
  & Variation in which the follower, trained on the left/right transfer areas, cannot pick up from the center.
  & Same as \scenario{Warehouse}; placing at the center transfer area is infeasible for the follower.
  & Pick up only from the left or right transfer area; central pick-up is impossible. \\
\hline
\addlinespace
\scenario{Warehouse (Pickup-unseen)}
  & Variation in which the follower, trained on the left/right transfer areas, can in principle pick up from any of the three.
  & Same as \scenario{Warehouse}; all three drop-offs are feasible.
  & Pick up from any transfer area; the central one was unseen during training. \\
  \addlinespace
\hline
\hline
\addlinespace
\scenario{Household}
  & Two robots operate in the same domestic environment.
  & Clean up dust under the table. The table has to be moved to reach it.
  & Traverse the narrow pathway and reach the opposite side. \\

\bottomrule
\end{tabular}
}
\caption{Overview of the evaluation scenarios. The \scenario{Warehouse} family shares the same physical environment but differs in the follower's training distribution and capability. \scenario{Household} is a hardware-only scenario.}
\label{tab:scenarios}
\end{table}

\textbf{Baselines}
To the best of our knowledge, \scenario{PAVE} is the only method that directly addresses the problem of sequential task continuation. Cooperative Consecutive Policies (CCP)~\cite{erskine_developing_2022} train temporally consistent consecutive policies for long-horizon tasks on a single robot by modifying the current agent’s policy to maximize the value estimated by the next agent’s critic. Although CCP is structurally different (both agents share the same embodiment, as well as the state and action spaces, and are trained jointly), CCP therefore serves as inspiration for our \baseline{Finetuning} baseline. Additionally, we perform a set of ablations to analyze the contribution of each component of \baseline{PAVE}.

\textbf{\baseline{Naive}.} The prior leader policy is rolled out without considering the follower's benefit. It provides a lower bound on the performance of our proposed approach and serves as a metric for determining whether guidance is possible at all.

\textbf{\baseline{PAVE (w/o OOD)}.} The proposed planner without the uncertainty filter. This serves as an ablation to assess the effectiveness of the mask in preventing the value function from being trusted in untrained regions of the state space.

\textbf{\baseline{PAVE (w/o $\pi_L$)}.} The proposed planner with an uninformed prior policy. This serves as an ablation to assess the effectiveness of using the leader's prior policy to guide the planning process.

\textbf{\baseline{Leader-only}.} The proposed planner without including the following value function in the optimization. This serves as an ablation to assess whether performance gains stem from online planning alone or from the use of the value function. 

\textbf{\baseline{Finetuning (FT)}.} Prior leader policy fine-tuned on the followers' return as a terminal reward computed by a rollout of the followers' policy from the handoff state. Because it relies on collective interaction data and further pre-deployment training (both of which \baseline{PAVE} forgoes), it serves as a privileged comparison.

\section{Results}
\begin{figure}[t]
	
    \centering
    \begin{adjustbox}{minipage=1\textwidth, center}
    \includegraphics[width=1.0\textwidth]{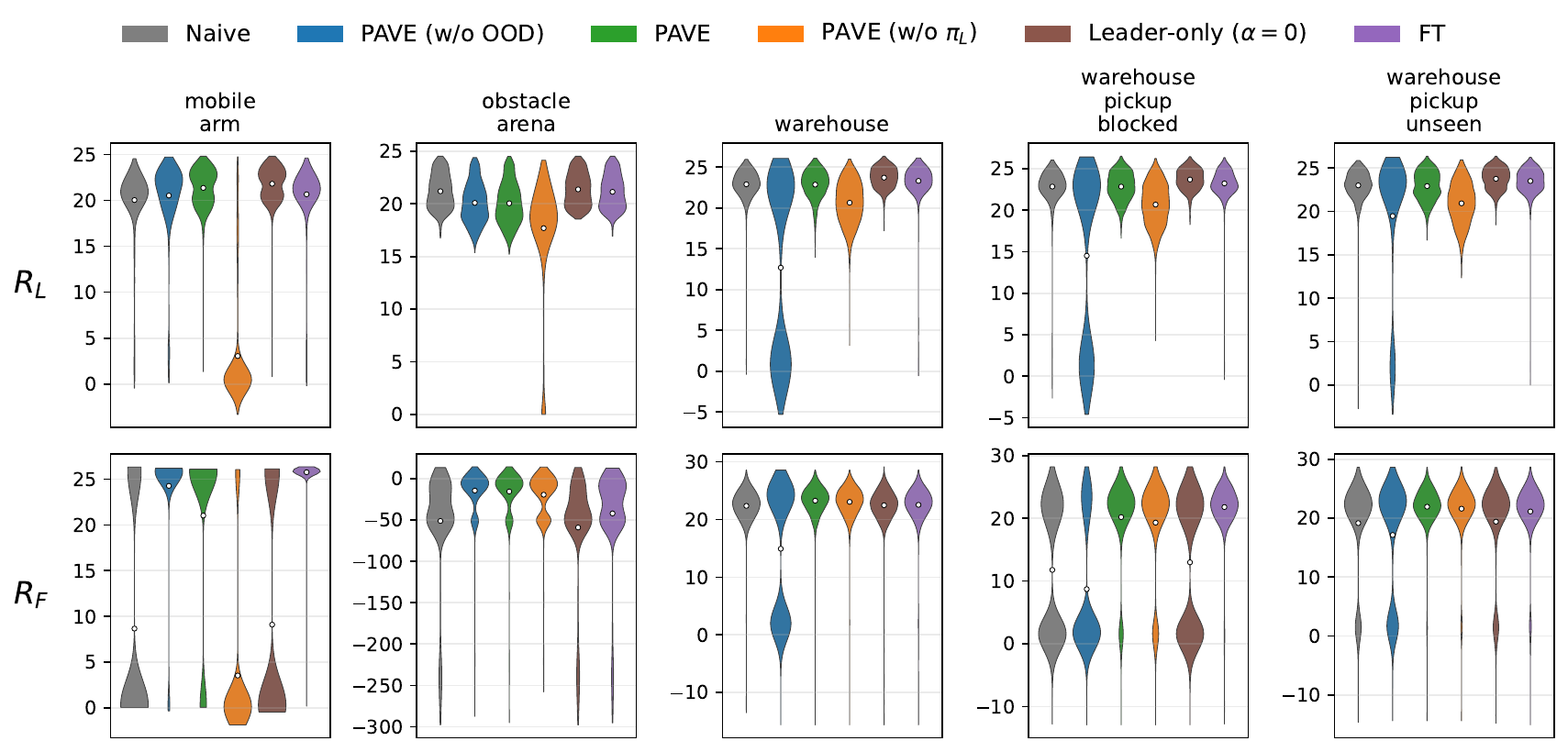}
    \end{adjustbox}
    \caption{Distribution Plots of Leader and Follower Return under baselines and our method. Leader behavior either supports or hinders the follower, making the followers' return distribution inherently multi-modal. Our approach successfully shifts this distribution toward the mode of follower success.}
    \label{fig:return_distributions}
\end{figure}

We evaluate our method against baselines in five simulated scenarios and one real-world deployment. Simulation results are obtained using six random seeds to compute the standard deviation. Figure \ref{fig:return_distributions} shows the return distributions, while Table \ref{tab:cross_scenario_success_rates} reports per-scenario success rates for both the leader and the follower, as well as their joint success.

\textbf{PAVE improves follower outcomes without sacrificing leader success.} We observe an average increase of \paveNaiveSFMean and a maximum increase of \paveNaiveSFMax in follower success rates across all scenarios. While the follower’s success rate in the \scenario{Warehouse} scenario is saturated, we still observe a small improvement in the follower’s return. The leader’s success rate is generally equal to or higher than that of the \baseline{NAIVE} baseline. The return distributions mirror the increase in follower success rates. We observe that the follower’s return distributions are typically multimodal, with each mode corresponding to either the success or failure outcome. \baseline{PAVE} shifts the return distribution toward the success mode, with higher returns.

\textbf{PAVE is robust to unseen handoff states.} The \scenario{Warehouse} family demonstrates the robustness of our approach to unseen handoff states. In both \scenario{Warehouse Pickup Blocked} and \scenario{Warehouse Pickup Unseen}, the follower’s value function overestimates the return of the object placed at the central gap as it interpolates between the two seen gaps. Further it wrongly extrapolates the handover value outside of transfer areas, assigning high value to areas where no handoff is possible. By employing the uncertainty filter and rejecting the misleading value estimates, \baseline{PAVE} is robust to this overestimation and significantly outperforms the \baseline{NAIVE} baseline. The $\baseline{PAVE (w/o OOD)}$ ablation underperforms \baseline{NAIVE} and $\baseline{PAVE}$ significantly, as it is led to the wrong gap or no gap at all, demonstrating that uncertainty estimation is essential for a successful handoff.

\textbf{Using an OOD filter trades off residual performance with catastrophic failures.}
\baseline{PAVE} trades residual performance in familiar regions for protection against catastrophic OOD failures. The filter is deliberately conservative: while it successfully blocks over-optimistic estimates in the \scenario{Warehouse} family, it also discards helpful smooth gradients where extrapolation would be informative. For example, in \scenario{mobile arm} the arm only observed objects in a cone in front of it during training, but the value function still assigns higher values smoothly to the general proximity of the base. \baseline{PAVE (w/o OOD)} can follow this incidentally useful value estimate to the optimal handoff orientations.

\textbf{Using the leader's prior as a planning initialization never hurts and helps substantially on constrained tasks.} \baseline{PAVE} generally outperforms \baseline{PAVE (w/o $\pi_L$)}. \scenario{Mobile Arm} and \scenario{Obstacle Arena} benefit the most from the initial guess. Empirically, we conclude that using the leader’s prior policy as an initial guess for the optimization often improves overall performance and never impairs it.

\textbf{PAVE matches finetuning without privileged information or pre-deployment training.} \baseline{PAVE} attains performance comparable to or better than the finetuned leader in three of five scenarios and trails it in the other two. Although joint training with shared critics is typically beneficial in multi-agent settings with synchronously acting agents, the sequential, long-horizon nature of task continuation makes it difficult for reinforcement learning alone to reliably converge to the optimal solution.

\textbf{Follower gains stem from value exchange and not from online planning alone.}
\baseline{Leader-only} improves leader success by up to $7.1$ pp but yields no consistent follower gain, recovering at most $14.2\%$ of \baseline{PAVE}'s follower improvements on the four scenarios where \baseline{PAVE} has headroom. Therefore, \baseline{PAVE}'s follower gains stem primarily from value exchange, not online planning. 

\textbf{PAVE transfers from simulation to a physical household setup.} In $20$ deployments with randomized initial conditions, we compare the performance of the \baseline{NAIVE} baseline with our PAVE method, running online in the $\scenario{Household}$ scenario (Figure \ref{fig:hero}). With both \baseline{NAIVE} and \baseline{PAVE}, the leader succeeds in every instance; however, \baseline{PAVE} increases the follower’s success rate by $40$~pp. The real collision dynamics differ substantially from those in simulation (which assume no friction and simple velocity matching), highlighting the strength of online replanning in compensating for prediction error.
\begin{table}[t]
\centering
\label{tab:cross_scenario_success_rates}
\resizebox{\textwidth}{!}{%
\begin{tabular}{ll|ccccc|c}
\toprule
Scenario &  & Naive & PAVE (w/o OOD) & PAVE & PAVE (w/o $\pi_L$) & Leader-only ($\alpha=0$) & FT \\
\midrule
\scenario{mobile arm} & $S_L=$ & 92.89\,$\pm$\,5.79 & 95.44\,$\pm$\,2.92 & 99.72\,$\pm$\,0.36 & 12.89\,$\pm$\,0.57 & 99.94\,$\pm$\,0.12 & 95.22\,$\pm$\,2.73 \\
 & $S_F=$ & 32.67\,$\pm$\,17.84 & 93.83\,$\pm$\,2.60 & 81.17\,$\pm$\,5.53 & 13.39\,$\pm$\,1.24 & 34.44\,$\pm$\,4.78 & 99.50\,$\pm$\,0.42 \\
 & $S_{L\wedge F}=$ & 30.28\,$\pm$\,15.81 & \textbf{93.61\,$\pm$\,2.81} & 81.00\,$\pm$\,5.31 & 10.39\,$\pm$\,0.68 & 34.44\,$\pm$\,4.78 & \textbf{95.11\,$\pm$\,2.73} \\
\midrule
\scenario{obstacle arena} & $S_L=$ & 100.00\,$\pm$\,0.00 & 100.00\,$\pm$\,0.00 & 100.00\,$\pm$\,0.00 & 94.22\,$\pm$\,4.72 & 100.00\,$\pm$\,0.00 & 100.00\,$\pm$\,0.00 \\
 & $S_F=$ & 45.78\,$\pm$\,11.71 & 84.67\,$\pm$\,6.43 & 83.61\,$\pm$\,6.64 & 74.17\,$\pm$\,7.40 & 36.39\,$\pm$\,11.12 & 54.83\,$\pm$\,12.62 \\
 & $S_{L\wedge F}=$ & 45.78\,$\pm$\,11.71 & \textbf{84.67\,$\pm$\,6.43} & 83.61\,$\pm$\,6.64 & 73.22\,$\pm$\,6.72 & 36.39\,$\pm$\,11.12 & 54.83\,$\pm$\,12.62 \\
\midrule
\scenario{warehouse} & $S_L=$ & 99.56\,$\pm$\,0.60 & 55.83\,$\pm$\,29.20 & 100.00\,$\pm$\,0.00 & 99.89\,$\pm$\,0.25 & 100.00\,$\pm$\,0.00 & 99.22\,$\pm$\,0.92 \\
 & $S_F=$ & 99.33\,$\pm$\,0.47 & 58.33\,$\pm$\,26.71 & 99.78\,$\pm$\,0.25 & 99.61\,$\pm$\,0.36 & 99.72\,$\pm$\,0.23 & 98.72\,$\pm$\,1.15 \\
 & $S_{L\wedge F}=$ & 99.11\,$\pm$\,0.83 & 55.67\,$\pm$\,29.18 & \textbf{99.78\,$\pm$\,0.25} & 99.61\,$\pm$\,0.36 & 99.72\,$\pm$\,0.23 & 98.50\,$\pm$\,1.27 \\
\midrule
\scenario{warehouse pickup blocked} & $S_L=$ & 99.28\,$\pm$\,1.08 & 62.56\,$\pm$\,14.58 & 100.00\,$\pm$\,0.00 & 99.94\,$\pm$\,0.12 & 100.00\,$\pm$\,0.00 & 99.72\,$\pm$\,0.30 \\
 & $S_F=$ & 49.78\,$\pm$\,10.26 & 32.50\,$\pm$\,7.01 & 89.94\,$\pm$\,10.02 & 85.67\,$\pm$\,1.53 & 55.50\,$\pm$\,7.65 & 98.11\,$\pm$\,2.02 \\
 & $S_{L\wedge F}=$ & 49.44\,$\pm$\,10.21 & 31.50\,$\pm$\,7.37 & \textbf{89.94\,$\pm$\,10.02} & 85.67\,$\pm$\,1.53 & 55.50\,$\pm$\,7.65 & \textbf{98.00\,$\pm$\,1.95} \\
\midrule
\scenario{warehouse pickup unseen} & $S_L=$ & 99.89\,$\pm$\,0.25 & 83.06\,$\pm$\,10.13 & 100.00\,$\pm$\,0.00 & 100.00\,$\pm$\,0.00 & 100.00\,$\pm$\,0.00 & 99.89\,$\pm$\,0.16 \\
 & $S_F=$ & 85.44\,$\pm$\,5.03 & 73.89\,$\pm$\,13.87 & 98.22\,$\pm$\,0.76 & 96.56\,$\pm$\,1.42 & 86.61\,$\pm$\,5.54 & 95.00\,$\pm$\,1.66 \\
 & $S_{L\wedge F}=$ & 85.44\,$\pm$\,5.03 & 73.67\,$\pm$\,14.02 & \textbf{98.22\,$\pm$\,0.76} & 96.56\,$\pm$\,1.42 & 86.61\,$\pm$\,5.54 & 95.00\,$\pm$\,1.66 \\
 \midrule
 \midrule
 \scenario{Household} & $S_L=$ & 100.00 & --& 100.00 & -- & --& -- \\
 (on hardware)& $S_F=$ & 30.00 & -- & 70.00 & --& --& -- \\
 & $S_{L \wedge F}=$ & 30.00 & --& 70.00 & --& -- & --\\
\bottomrule
\end{tabular}
}
\caption{Leader / follower / joint success rates (\%) per scenario across modes, each as mean $\pm$ std across $6$ seeds. $S_L$ / $S_F$ = leader/follower success rate. $S_L \wedge S_F$ is computed per seed (counting the rollouts in which both leader and follower succeeded) and then aggregated across seeds.
\textbf{Bold} marks the maximum between methods without access to privileged information. \baseline{FT} serves as a privileged comparison; \textbf{bold} when it outperforms \baseline{PAVE}.
\scenario{Household} is hardware-only and reports the success rates of $20$ leader-follower deployments with randomized initial conditions.}
\label{tab:cross_scenario_success_rates}
\end{table}

\section{Conclusion}
\label{sec:conclusion}
Extrapolating from current developments, robots will soon be ubiquitous in daily life. While current research often focuses on building optimal systems for individual tasks, the growing number of robots in shared spaces will require simple and efficient methods and protocols to coordinate heterogeneous robotic systems. In this paper, we took a step towards such an interface and proposed the use of the environmental value function to serve as such a protocol.
We have shown that an \textit{agent's value function}, when paired with an \textit{uncertainty filter}, serves as an effective interface between learning-based agents. We demonstrate that our method improves follower success rates by \paveNaiveSFMean on average and by up to \paveNaiveSFMax, with the greatest gains occurring when the leader's selfish choices are most detrimental to the follower's success. This suggests that the value-function interface is most effective when cooperation matters most.
Uncertainty filtering through out-of-distribution detection is necessary to prevent catastrophic failures. Without it, the leader may follow the follower's value function into unseen environmental states, which degrades performance.

\textbf{Limitations and Future Work}
This work focuses on \textit{two agents acting sequentially in a pre-defined order}. 
One axis of future research lies in the number of possible followers. If the sequence of agents acting in the environment is either unknown or not predetermined, one leader could adapt to the needs of multiple followers; a possible solution is optimizing a weighted combination of their value functions or lexicographically optimizing in an order of priority. 
A critical direction for future work is to tackle nested optimization as a second axis of scaling. By extending our planning horizon beyond the immediate follower, future models can optimize for complex, multi-step sequences involving multiple downstream agents.
Due to runtime constraints of online adaptation, the planning horizon is limited. As a result, the handoff state has to be approximated with the state prediction at the end of the planning horizon. This issue is partly alleviated by the uncertainty filter, since mid-leader-episode handoff states are often not observed by the follower, reducing the influence of intermediate approximations until the end of the episode approaches. Nonetheless, this remains a critical challenge when considering environment modifications that require a longer acting horizon.
This work assumes that both agents \textit{``share a common representation of the environment $\mathbf{w}_T$''}. Because robots often serve different purposes, their world representations may also be task-specific. Future work on either transformations between different representations or a general shared world representation could help extend this framework.
Furthermore, we assume that the follower shares a truthful value function and uncertainty estimate with the leader. A malicious follower could mislead the leader to produce unfavorable or even dangerous environmental states, making further research into approaches to increase safety necessary.

\newpage

\clearpage

\acknowledgments{This work is supported by European Research Council (ERC) Project 949940 (gAIa).}

\bibliography{references}  %

\newpage
\appendix

\section{Symbols}
\label{apL}

\begin{table}[H]
\centering

\label{tab:notation}
\begin{tabular}{@{}cl@{}}
\toprule
\textbf{Symbol} & \textbf{Description} \\
\midrule

\multicolumn{2}{@{}l}{\textit{Agents \& Policies}} \\[2pt]
$\pi_L$       & Leader's policy \\
$\pi_F$       & Follower's policy (fixed, independently trained) \\
$r_L, r_F$    & Leader's and follower's reward functions \\[6pt]

\multicolumn{2}{@{}l}{\textit{States, Distributions, and Dynamics}} \\[2pt]
$\mathbf{w}$              & Shared environment state \\
$\mathbf{s}^L, \mathbf{s}^F$ & Leader's and follower's private states \\
$\mathbf{w}_t$            & Environment state at time $t$ \\
$\mathbf{s}^L_t, \mathbf{s}^F_t$ & Leader's and follower's private states at time $t$ \\
$\mathbf{w}_T$            & Terminal environment state (handoff state from leader to follower) \\
$\mathbf{w}_{t+H}$        & Environment state at the end of the planning horizon \\
$\rho_L, \rho_F$          & Initial distributions of leader's and follower's private states \\
$\rho_W$                  & Initial distribution of the environment state \\
$H_T(\pi_L)$              & Handoff distribution induced by leader policy $\pi_L$ over $T$ steps \\
$T$                       & Leader's episode length \\
$\gamma$                  & Discount factor \\[6pt]

\multicolumn{2}{@{}l}{\textit{Returns and Value Functions}} \\[2pt]
$J_L(\pi_L)$              & Leader's expected return \\
$J_F(\pi_F, \mathbf{w})$  & Follower's expected return from handoff state $\mathbf{w}$ \\
$V_F(\mathbf{w})$         & Environmental value function of the follower \\
$V_F^{\mathrm{critic}}(\mathbf{s}, \mathbf{w})$ & PPO-trained follower critic (depends on private state) \\
$V_L^{\mathrm{critic}}(\mathbf{s}, \mathbf{w})$ & Leader's value function (used for tail estimation in planning) \\
$\alpha$                  & Cooperation weighting factor \\
$J_L^{\min}, \Delta J_L$  & Leader return normalization constants \\
$J_F^{\min}, \Delta J_F$  & Follower value normalization constants \\

\multicolumn{2}{@{}l}{\textit{OOD Uncertainty Filter}} \\[2pt]
$\beta(\mathbf{w}_T)$     & KNN-based OOD score, $\in [0, 1]$ \\
$K$                       & Number of nearest neighbors \\
$\mathtt{neighbor}(\mathbf{w}_T, k)$ & $k$-th nearest neighbor of $\mathbf{w}_T$ in the training distribution \\
$\tau$                    & Distance threshold below which states are considered in-distribution \\
$d_{\max}$                & Distance ceiling above which states are fully out-of-distribution \\[6pt]

\multicolumn{2}{@{}l}{\textit{Planning}} \\[2pt]
$H$                       & Planning horizon (receding-horizon) \\
$r^L_i$                   & Leader's reward at time $i$ \\
$\mathbf{s}_i, \mathbf{w}_i$ & Predicted states within the planning horizon \\

\bottomrule
\end{tabular}
\end{table}

\newpage

\section{Extended Related Work}
\label{ap:extended_related_work}

\paragraph{Ad Hoc Teamwork}

Stone et al.~\cite{stone_ad_2010} define the challenge of creating ad hoc teams of autonomous agents as: \emph{``To create an autonomous agent that is able to efficiently and robustly collaborate with previously unknown teammates on tasks to which they are all individually capable of contributing as team members.''}
PLASTIC~\cite{barrett_making_2017} addresses the problem in two variants. PLASTIC-Model learns teammate models from prior teams and plans against them at deployment, while PLASTIC-Policy learns a library of cooperative policies and selects the best-fitting one online based on the observed teammate behavior.
TEAMSTER~\cite{ribeiro_teamster_2023} builds on ideas in PLASTIC and not only learns a model of the teammates, but also a model of the environment. For a detailed survey of Ad Hoc Teamwork, the reader is referred to \cite{mirsky_survey_2022}.
Our approach differs from Ad Hoc Teamwork in that it explicitly allows direct communication between agents, whereas the Ad Hoc Teamwork literature typically omits such communication. We further consider agents that perform distinct tasks rather than receive a shared payoff.
\paragraph{Zero Shot Coordination}
Zero Shot Coordination makes different assumptions than Ad Hoc Teamwork, as it considers games where players must choose and fix their strategies prior to being deployed together. Unlike Ad Hoc Teamwork, this does not allow for online adaptation to teammates, but requires fixed policies that are robust to a range of different partners. Other-Play \cite{hu_other-play_2020} proposes a training framework that trains policies to be robust to independently trained agents using equivalence mappings, thereby preventing task-specific conventions arising from Self-Play. As PAVE specifically assumes message-passing and adaptation to the previously unseen partner, it is not a Zero-Shot Coordination method, but addresses a related problem.
\paragraph{Multi-Agent Reinforcement Learning and Shared Critics}
Multi-Agent Reinforcement Learning extends classical Single-Agent Reinforcement Learning to Decentralized POMDPs in which multiple agents jointly act in the same environment. MARL algorithms are typically classified by whether they are centralized (C) or decentralized (D) at training (T) and execution (E) time. The CTCE paradigm considers all actors as a single agent employing single-agent RL on their joint state and action space. These approaches typically do not scale well with the number of agents. Approaches like IPPO~\cite{de2020independent} and I2Q~\cite{jiang_i2q_2022} belong to the DTDE paradigm and consider each agent as an individual learner interacting with the environment at the same time. While suffering less from poor scaling in the number of agents, multiple agents learning in the same environment introduce non-stationarity in the environmental dynamics from the perspective of any single agent. CTDE approaches like MAPPO~\cite{yu_surprising_2022}, MADDPG~\cite{lowe_multi-agent_2017}, COMA~\cite{foerster_counterfactual_2018}, and QMIX~\cite{rashid_qmix_2018} introduce a shared critic function to the training process, targeting non-stationary and credit assignment problems to stabilize the training process.
	
MARL typically does not consider long-horizon sequential tasks in which one agent acts for an entire episode before handing off the environment to a second agent. In MARL, sequentiality typically refers to the sequential execution of a single action at one timestep and appears in the context of credit assignment. Zang et al.~\cite{zang2023sequential} propose a sequential MARL framework in which the value function is factorized according to the sequence of agents acting on the environment.
Erskine et al.~\cite{erskine_developing_2022} consider sequential tasks with handoff, but restrict their setting to multiple policies acting on the same robot. Because the embodiment remains unchanged, achieving consistency between different agents is ensured by training the agents with a mixture of subsequent Q-functions.
In contrast, our approach considers agents with different embodiments, distinct action spaces, and modifiable environments.

\paragraph{Stigmergy}
Stigmergy is a natural process that organizes the behavior of animal swarms \cite{dorigo_ant_2000,theraulaz_brief_1999,holland_stigmergy_1999}. Ants, for example, leave pheromone trails while foraging, which other ants can follow to a food source. Over time, the paths that individuals take provably converge to optimal solutions, which is why stigmergy has inspired the class of Ant Colony Optimization (ACO) algorithms \cite{dorigo_ant_2006}. Although our proposed approach also considers modifications to the environment to influence other agents' behavior, it directly optimizes the follower's return instead of using the environment as an implicit communication variable. Therefore, while our approach shares some of the characteristics of stigmergy, it is not strictly stigmergic.

\paragraph{Environmental Co-Optimization}
As we consider sequential scenarios where the agents cooperate solely through a shared environment, our work connects to prior work on environmental optimization. Gao et al.~\cite{gao_constrained_2023, gao_co-optimizing_2025}, and Li et al.~\cite{li_scaling_2025} consider the environment as a decision variable together with the policies of agents acting within it and jointly optimize both in a bi-level iterative optimization scheme.  While our work also considers the effects of a given environment on the return and task success of a given agent, we clearly differentiate from existing literature in two ways. Firstly, we do not consider the environment as an explicit decision variable, as it can only be modified indirectly through the leading agent's behavior, and secondly, we assume that the second agent's policy does not adapt to the changing environment. 

\paragraph{Sampling-Based Model Predictive Control with Learned Cost Functions}
PAVE uses a sampling-based MPC formulation~\cite{williams2017model} with the Cross-Entropy Method~\cite{rubinstein_cross-entropy_1999} as the inner optimizer, similar to its use in model-based RL planners such as PETS~\cite{chua2018deep}; recent CEM variants such as iCEM~\cite{pmlr-v155-pinneri21a} are complementary and could be incorporated.
For tail estimation, PAVE bootstraps the leader's return using its own value function $V_L$, following standard practice in 
value-augmented MPC~\cite{lowrey2018plan, Hansen2022tdmpc}. The general principle of pairing a search procedure with a learned value estimate was demonstrated at scale by AlphaGo Zero \cite{silver2017mastering}, which used a learned value function to evaluate leaf nodes within Monte Carlo Tree Search.
The role of the follower's value function $V_F$ differs as it serves as the coordination interface through which the two independently trained agents communicate their preferences.
PAVE additionally employs a $k$-nearest-neighbor distance-based Out-Of-Distribution detector~\cite{sun_out--distribution_2022, yang2024generalized} to down-weight the value-function term in environment state regions in which the follower critic has not been trained to prevent inaccurate value approximations.

\paragraph{Learned Communication}
Multi-agent learning has been shown to benefit from learned communication channels~\cite{foerster_learning_2016, li_graph_2020, li2021message},  where messages emerge from joint training and condition agents' policies on one another at every timestep. PAVE differs in two ways. First, communication is one-shot: the follower transmits its value function once, and the leader plans and acts without further exchange. Second, the transmitted message has a known meaning (the follower's expected return from a given environment state), instead of being a task- and training-team-specific representation, only interpretable between co-trained agents. PAVE's message is learned but interpretable, enabling pairing of arbitrary agents that have a shareable value function.

\newpage

\section{Simulation Environments}
\label{ap:simulation_environments}
All simulation environments are custom and written in numpy. The following explains the rewards structure of individual environments and lists their numerical values. Figure \ref{fig:scenario_cards} depicts \scenario{Obstacle Arena}, \scenario{Mobile Arm} and \scenario{Warehouse}. Figure \ref{fig:hero} and Figure \ref{fig:arena_raw} show the \scenario{Household} scenario.

\subsection{\scenario{Mobile Arm}}
The leader must steer the mobile base to a specified distance from the target object and then terminate the episode by outputting a termination signal from its action space. The follower controls a 4-DoF arm, consisting of the arm’s yaw and three parallel joints to pick up the object and place it in a tray on the mobile base. The object is picked up and placed when the end effector comes sufficiently close to the corresponding pick-up and placement points.

\begin{table}[H]
        \centering
        \caption{\scenario{mobile\_arm} environment parameters. }
        \begin{minipage}[t]{0.48\textwidth}
            \centering
            \textbf{Leader} \\[0.4em]
            
\begin{tabular}{lc}
\toprule
Parameter & Value \\
\midrule
\texttt{step\_penalty} & 0.01 \\
\texttt{progress\_weight} & 1 \\
\texttt{goal\_reward} & 25 \\
\texttt{gaussian\_sigma} & 0.3 \\
\bottomrule
\end{tabular}

        \end{minipage}\hfill
        \begin{minipage}[t]{0.48\textwidth}
            \centering
            \textbf{Follower} \\[0.4em]
            \begin{tabular}{lc}
\toprule
Parameter & Value \\
\midrule
\texttt{step\_penalty} & 0.01 \\
\texttt{progress\_weight} & 1 \\
\texttt{pickup\_reward} & 1 \\
\texttt{goal\_reward} & 25 \\
\texttt{collision\_penalty} & 0 \\
\bottomrule
\end{tabular}

        \end{minipage}
    \end{table}%

\subsection{\scenario{Obstacle Arena}}
The leader has to clear the lower half of the arena by pushing the obstacle out of the center and into contact with any wall; a follower must then traverse the room by navigating through a narrow pathway and exiting it at the lower boundary. Their objectives conflict when the leader pushes the obstacle into contact with the central wall, blocking the passageway.

    \begin{table}[H]
        \centering
        \caption{\scenario{obstacle\_arena} environment parameters. }
        \begin{minipage}[t]{0.48\textwidth}
            \centering
            \textbf{Leader} \\[0.4em]
            
\begin{tabular}{lc}
\toprule
Parameter & Value \\
\midrule
\texttt{goal\_reward} & 25 \\
\bottomrule
\end{tabular}

        \end{minipage}\hfill
        \begin{minipage}[t]{0.48\textwidth}
            \centering
            \textbf{Follower} \\[0.4em]
        
\begin{tabular}{lc}
\toprule
Parameter & Value \\
\midrule
\texttt{goal\_reward} & 25 \\
\texttt{collision\_penalty} & 300 \\
\texttt{progress\_weight} & 1 \\
\texttt{distance\_weight} & 0.1 \\
\bottomrule
\end{tabular}

        \end{minipage}
    \end{table}%

\subsection{\scenario{WAREHOUSE} Family} 
\baseline{PAVE} is evaluated on three variations of the \scenario{WAREHOUSE} with the same underlying reward structure. The leader is tasked with picking up an object in the lower half of the arena and transporting it to one of the three transfer areas. The follower has to navigate through the shelves to the transfer area, pick up the object, and place it at a goal position on the shelf. Pickup and placing occur automatically when the leader or follower is close enough to the object and oriented towards it.

    \begin{table}[H]
        \centering
        \caption{\scenario{warehouse} environment parameters. }
        \begin{minipage}[t]{0.48\textwidth}
            \centering
            \textbf{Leader} \\[0.4em]
            \begin{tabular}{lc}
\toprule
Parameter & Value \\
\midrule
\texttt{step\_penalty} & 0.01 \\
\texttt{step\_penalty\_carrying} & 0.05 \\
\texttt{pickup\_reward} & 1 \\
\texttt{goal\_reward} & 25 \\
\texttt{progress\_weight} & 1 \\
\bottomrule
\end{tabular}

        \end{minipage}\hfill
        \begin{minipage}[t]{0.48\textwidth}
            \centering
            \textbf{Follower} \\[0.4em]
        \begin{tabular}{lc}
\toprule
Parameter & Value \\
\midrule
\texttt{step\_penalty} & 0.01 \\
\texttt{step\_penalty\_carrying} & 0.1 \\
\texttt{pickup\_reward} & 4 \\
\texttt{goal\_reward} & 25 \\
\texttt{progress\_weight} & 1 \\
\texttt{collision\_penalty} & 0.1 \\
\bottomrule
\end{tabular}

        \end{minipage}
    \end{table}%

\subsection{\scenario{HOUSEHOLD}}
This scenario is similar to \scenario{Obstacle Arena} and only differs in the leader's objective. Instead of pushing the obstacle into contact with the wall, it has to move the obstacle out of the way to reach the dirt that is positioned under the table.

    \begin{table}[H]
        \centering
        \caption{\scenario{household} environment parameters. }
        \begin{minipage}[t]{0.48\textwidth}
            \centering
            \textbf{Leader} \\[0.4em]
            
\begin{tabular}{lc}
\toprule
Parameter & Value \\
\midrule
\texttt{step\_penalty} & 0.01 \\
\texttt{progress\_weight} & 3 \\
\texttt{wall\_collision\_penalty} & 0.1 \\
\texttt{pickup\_reward} & 25 \\
\bottomrule
\end{tabular}

        \end{minipage}\hfill
        \begin{minipage}[t]{0.48\textwidth}
            \centering
            \textbf{Follower} \\[0.4em]
        \begin{tabular}{lc}
\toprule
Parameter & Value \\
\midrule
\texttt{progress\_weight} & 3 \\
\texttt{goal\_reward} & 25 \\
\texttt{collision\_penalty} & 10 \\
\texttt{distance\_weight} & 0.1 \\
\bottomrule
\end{tabular}

        \end{minipage}
    \end{table}%

\section{Training and Model Selection}
\label{ap:training}
All leaders and followers across all scenarios use policy networks with two hidden layers of size 64 each, with a Gaussian policy head (SB3 defaults). Table \ref{tab:training_budget} shows the training budget for the leader and follower per scenario. The training steps were increased until policy convergence was reached.

Table \ref{tab:cross_scenario_success_rates} was generated by evaluating our method and the baseline over six independent training and evaluation seeds. 
\baseline{PAVE} assumes capable actors, so we discard policies with under $70\%$ individual success. Applied uniformly across all baselines, this ensures a fair cross-method comparison. This removes noise from followers that fail regardless of interaction method while preserving variance among competent policies. No further selection of seeds was performed based on the performance of our method or baselines. 

\begin{table}[H]
\centering
\caption{PPO (Stable-Baselines3, \texttt{MlpPolicy}) algorithm hyperparameters.}
\label{tab:ppo_defaults}
\begin{tabular}{lc}
\toprule
Hyperparameter & Value \\
\midrule
Learning rate & 0.0003 \\
Rollout length $n_{\text{steps}}$ (per env) & 2048 \\
Minibatch size & 64 \\
Epochs per update & 10 \\
Discount $\gamma$ & 0.99 \\
GAE $\lambda$ & 0.95 \\
PPO clip range $\epsilon$ & 0.2 \\
Entropy coef.\ $c_{\text{ent}}$ & 0 \\
Value-function coef.\ $c_{vf}$ & 0.5 \\
Max gradient norm & 0.5 \\
\bottomrule
\end{tabular}
\end{table}

\begin{table}[H]
\centering
\caption{Per-agent training budget: Follower/leader total environment steps}
\label{tab:training_budget}

\begin{tabular}{lcc}
\toprule
Scenario & Follower\ steps &  Leader\ steps  \\
\midrule
\scenario{mobile\_arm} & 2\,M &  1\,M  \\
\scenario{obstacle\_arena} & 500\,k & 400\,k \\
\scenario{warehouse} & 1.5\,M &  500\,k  \\
\scenario{warehouse\_pickup\_blocked} & 1.5\,M & 500\,k  \\
\scenario{warehouse\_pickup\_unseen} & 1.5\,M &  500\,k  \\
\scenario{household} & 1.5\,M & 400\,k  \\
\bottomrule
\end{tabular}

\end{table}

All \baseline{Finetuning} experiments were run for an additional 1.5M steps. The return signal is computed as the sum of the normalized leader return and the normalized follower return, with the latter scaled by $\alpha = 1$. This differs from the $\alpha = 4$ used in \baseline{PAVE}. We made this choice because \baseline{Finetuning} with $\alpha = 1$ yielded higher success rates than with $\alpha = 4$, so we report only the best-performing version of the \baseline{Finetuning} baseline.

\newpage
\section{Hardware Setup}
\label{ap:hw_setup}
\begin{figure}[H]
    \centering
    \includegraphics[width=0.8\linewidth]{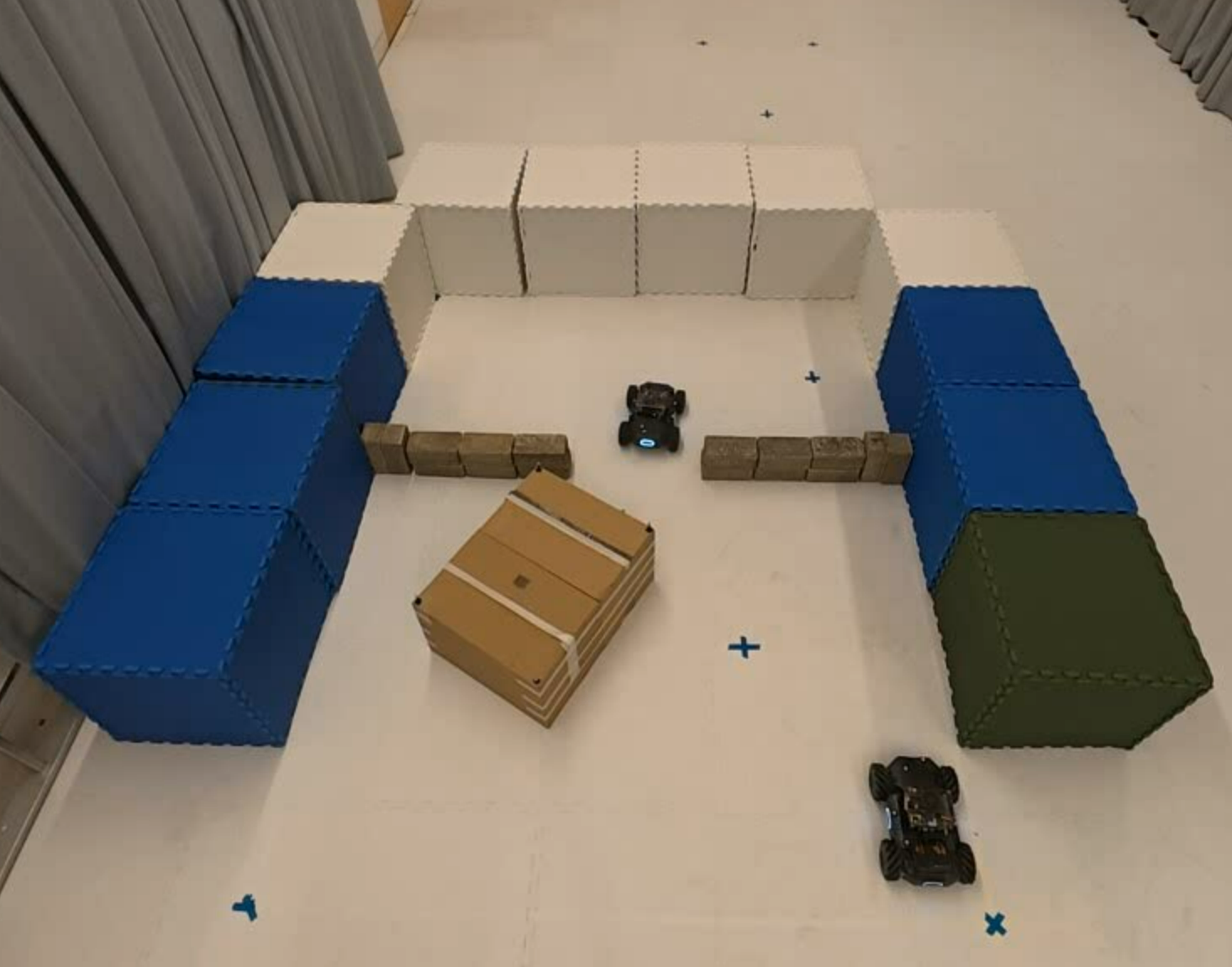}
    \caption{Hardware setup: Two RoboMaster robots and the table (box) in
the \scenario{Household} arena, tracked by OptiTrack. The dirt is
virtual and co-located with the initial table position. }
    \label{fig:arena_raw}
\end{figure}

For the deployment of the \scenario{Household} scenario, we used two RoboMaster \cite{blumenkamp2024robomaster} robots. The robots were velocity-controlled using the Freyja controller \cite{shankar2021freyja} at 50 Hz. Both the robots and the box representing the table in the scenario were tracked using OptiTrack, and the resulting pose data were used as feedback for the policies at 50 Hz. The dirt was purely virtual and initialized at the table's initial position.
Policies were trained in simulation and deployed on the real robots without further modification. \baseline{PAVE} used the nominal model from simulation to optimize trajectories and adapt the leader's policy. Figure \ref{fig:arena_raw} shows the experimental setup.
\paragraph{Trials} To compare \baseline{Naive} with \baseline{PAVE}, both methods were deployed across 20 trials with varied initial conditions; each trial used the same starting condition for \baseline{Naive} and \baseline{PAVE} so that the methods could be compared on identical configurations. Between trials, the initial conditions were randomized. The table was placed in the center of the lower half of the arena at different orientations, and the leader and follower were placed in the lower and upper halves of the arena, respectively.
\paragraph{Sim-To-Real Transfer and Failure Modes} In the absence of collisions, RoboMaster robots can be velocity-controlled with minimal tracking error, which is why the sim-to-real gap for the follower was negligible. In simulation, the interaction between the leader and the table is modeled using simple velocity matching, where the leader effectively imposes a translational and rotational velocity on the table depending on the contact angle. This resulted in a mismatch between simulation and real-world deployment, as the friction between the floor and the box is relatively high, but it did not necessarily diminish performance, since receding-horizon replanning could compensate for the model error. Nevertheless, real-world trajectories tended to be less smooth: the leader frequently required several push attempts within a single trial to successfully reposition the object, re-engaging when the initial push slipped or under-rotated.

\newpage

\newpage
\section{Planner Details}
\label{ap:planner_parameters}
The CEM planner uses a Gaussian action distribution with elite selection. We warm-start the action distributions by setting the mean to the actions obtained from rolling out the naive leader policy through the dynamics, and we initialize the covariance to $\sigma_\text{noise}$. At each iteration, we draw samples from the action distribution and evaluate them with respect to the objective defined in Equation \eqref{eq:pave_optim_problem}. We then select the best samples and update the action distribution by fitting a Gaussian to them. To prevent covariance collapse, we impose a lower bound on the covariance.

We instantiate the cost using the same parameters and procedures across all scenarios. Specifically, we use a follower value weighting term of $\alpha = 4$, a discount factor of $\gamma = 0.99$, and normalize the leader and follower returns to $[0,1]$ using the returns from 200 rollout episodes of the prior policies.

\begin{table}[H]
\centering
\caption{Cross-entropy-method (CEM) planner hyperparameters per scenario.}
\label{tab:cem_planner}
\resizebox{\textwidth}{!}{%
\begin{tabular}{lcccccccc}
\toprule
Scenario & Horizon & Samples & Iters & Elite frac. & $\sigma_{\text{noise}}$ & Min.\ std & Replan & Dist. \\
\midrule
\scenario{mobile\_arm} & 30 & 200 & 4 & 0.2 & 0.5 & 0.1 & 5 & gaussian \\
\scenario{obstacle\_arena} & 40 & 300 & 10 & 0.3 & 0.5 & 0.208 & 3 & gaussian \\
\scenario{warehouse} & 40 & 500 & 10 & 0.2 & 1.5 & 0.1 & 5 & gaussian \\
\scenario{warehouse\_pickup\_blocked} & 40 & 500 & 10 & 0.2 & 1.5 & 0.1 & 5 & gaussian \\
\scenario{warehouse\_pickup\_unseen} & 40 & 500 & 10 & 0.2 & 1.5 & 0.1 & 5 & gaussian \\
\scenario{household} & 20 & 100 & 5 & 0.3 & 0.5 & 0.2 & 5 & gaussian \\
\bottomrule
\end{tabular}
}
\end{table}

\newpage
\section{Case Study: Value Function and Uncertainty Estimate}
\begin{figure}[H]
    \centering
    \includegraphics[width=0.8\linewidth]{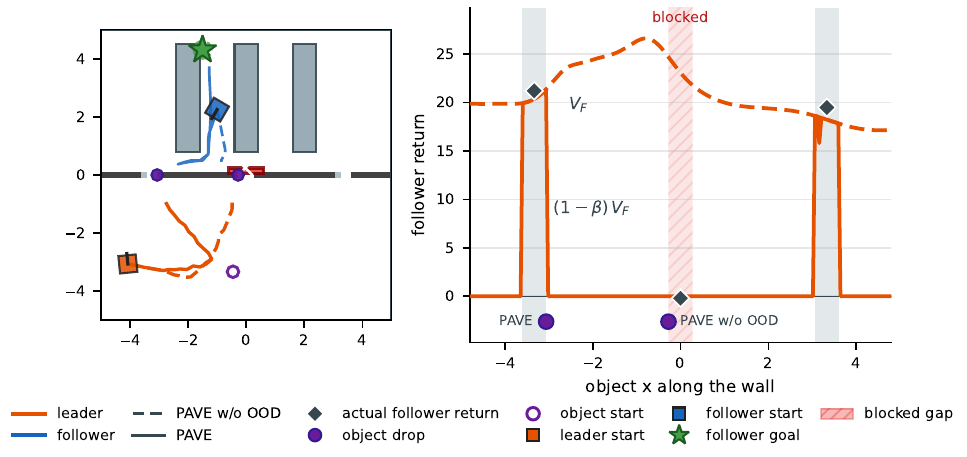}
    \caption{Comparison of trajectories  \baseline{PAVE (w/o OOD)} (dashed) and \baseline{PAVE} (solid) in one example of the \scenario{Warehouse Pickup Blocked} scenario. \textbf{Left}: Leader/Follower Trajectories. \textbf{Right:} Value function and uncertainty-weighted value function evaluated along the middle of the arena. Areas in which a drop is possible are marked in grey, blocked areas are marked in red.}
    \label{fig:warehouse_v_ood_trajectories}
\end{figure}

Figure \ref{fig:warehouse_v_ood_trajectories} shows the trajectories of the leader and follower policies for the \baseline{PAVE (w/o OOD)} baseline and \baseline{PAVE} in the \scenario{Warehouse Pickup Blocked} scenario, along with the value function, and uncertainty-weighted value function evaluated over the transfer area.
This example demonstrates the effectiveness of combining a value function with an uncertainty metric. The value function catastrophically overestimates the return of the central transfer area, even though pick-up is not possible from this location. Without an uncertainty estimate, \baseline{PAVE (w/o OOD)} is unaware that the follower has not been trained on experience involving the central drop-off area, and therefore trusts the estimate and drops the object in the center, leading to an adverse outcome for the follower.
In contrast, \baseline{PAVE} optimizes the uncertainty-weighted value function, disregarding the erroneous value estimate and instead selecting the next-best transfer area with low uncertainty.

\newpage

\section{Decomposition of the Leader Objective}
In this section, we call a leader policy \emph{selfish} if it optimizes only
the leader's individual objective, without regard to the follower. Formally,
given a feasible leader policy class \(\Pi_L\), a selfish leader policy is any
\[
\pi_L^\circ \in \arg\max_{\pi_L \in \Pi_L} J_L(\pi_L).
\]

For a cooperation weight \(\alpha \ge 0\), we define the
cooperative objective
\[
U_\alpha(\pi_L)
= J_L(\pi_L) + \alpha\, \E_{\mathbf{w} \sim H_T(\pi_L)}\!\left[J_F(\pi_F, \mathbf{w})\right].
\]
We call a leader policy \(U_\alpha\)-optimal if it maximizes
\(U_\alpha\) over \(\Pi_L\). For any alternative leader policy \(\pi'_L\), the improvement over the selfish leader decomposes as
\[
U_\alpha(\pi'_L)-U_\alpha(\pi_L^\circ)
=
\alpha
\big(
\E_{\mathbf{w} \sim H_T(\pi'_L)}\!\left[J_F(\pi_F, \mathbf{w}) \right] - \E_{\mathbf{w} \sim H_T(\pi_L^\circ)}\!\left[J_F(\pi_F, \mathbf{w}) \right ]
\big)
-
\big[
J_L(\pi_L^\circ)-J_L(\pi'_L)
\big].
\]
Consequently, for \(\alpha>0\), adapting the leader can improve the cooperative objective exactly when the induced follower gain exceeds the leader loss
scaled by \(1/\alpha\):
\[
\E_{\mathbf{w} \sim H_T(\pi'_L)}\!\left[J_F(\pi_F, \mathbf{w}) \right] - \E_{\mathbf{w} \sim H_T(\pi_L^\circ)}\!\left[J_F(\pi_F, \mathbf{w}) \right ]
>
\frac{
J_L(\pi_L^\circ)-J_L(\pi'_L)
}{\alpha}.
\]
For \(\alpha=0\), the cooperative objective reduces to the selfish leader
objective, so any selfish leader's policy is already optimal.

This identity also clarifies when no adaptation can improve over the selfish leader. If the selfish leader already produces a handoff state that is optimal for the follower,
i.e.,
\[
\E_{\mathbf{w} \sim H_T(\pi_L^\circ)}\!\left[J_F(\pi_F, \mathbf{w}) \right]
=
\sup_{\pi_L\in\Pi_L} \E_{\mathbf{w} \sim H_T(\pi_L)}\!\left[J_F(\pi_F, \mathbf{w}) \right], 
\]then no feasible deviation can improve the follower term, while every
deviation does not increase the leader term. In that case, the selfish
leader is already \(U_\alpha\)-optimal for every \(\alpha\ge 0\).

\section{Generative AI Usage Disclosure}
The authors used Claude (Anthropic) for language editing, writing suggestions, and LaTeX figure boilerplate, AI tools to explore related ideas and literature, and Claude Code for code generation. 
All content and code were reviewed by the authors, who take full responsibility for them.

\section{Reproducibility Statement}
Our method is described in Section~\ref{sec:planning}, with architectures, hyperparameters, training details, and OOD parameters in the Appendix. The complete code is publicly available at \url{https://github.com/proroklab/PAVE}.
\end{document}